%% file: main.tex
\documentclass[letterpaper, 10 pt, conference]{ieeeconf}
\IEEEoverridecommandlockouts
\usepackage{amsmath}
\usepackage{verbatim}
\usepackage{amssymb}
\usepackage{amsthm}
\usepackage{xcolor}
\usepackage{subcaption}
\usepackage[pdftex]{graphicx}
\usepackage{algorithm}
\usepackage{algorithmicx}
\usepackage{algpseudocode}
\usepackage{pgf}
\usepackage{pgfplots}
\usepackage{float}
\usepackage{listings}
\usepackage{xspace}
\usepackage{tabularx}
\usepackage{colortbl}
\usepackage{tikz}
\usepackage{multirow}
\usepackage{pifont}%
\usepackage{cuted}
\pgfplotsset{compat=1.18}
\algrenewcommand\algorithmicrequire{\textbf{Input:}}
\algrenewcommand\algorithmicensure{\textbf{Output:}}

\usepackage[
	activate   = {true},
	protrusion = false,
	expansion  = true,
	kerning    = true,
	spacing    = true,
	tracking   = false,
	auto       = true,
	selected   = true,
	factor     = 2000,
	stretch    = 30,
	shrink     = 30,
]{microtype}

\usepackage{csquotes}
\usepackage[
maxbibnames=99,
maxcitenames=2,
natbib=true,
style=numeric-comp,
backend=biber,
sorting=none,
giveninits=true,
url=false, 
doi=false,
eprint=false,
isbn=false,
]{biblatex}

\definecolor{mycolor}{RGB}{215, 25, 28}
\definecolor{TODO}{RGB}{34, 139, 34}
\definecolor{highlight}{RGB}{34, 139, 34}

\definecolor{codegreen}{rgb}{0,0.6,0}
\definecolor{codegray}{rgb}{0.5,0.5,0.5}
\definecolor{codepurple}{rgb}{0.58,0,0.82}
\definecolor{backcolour}{rgb}{0.95,0.95,0.92}

\definecolor{lightgrey}{rgb}{0.9,0.9,0.9}
\definecolor{grey}{rgb}{0.7,0.7,0.7}

\lstdefinestyle{mystyle}{
    backgroundcolor=\color{backcolour},   
    commentstyle=\color{codegreen},
    keywordstyle=\color{magenta},
    numberstyle=\tiny\color{codegray},
    stringstyle=\color{codepurple},
    basicstyle=\ttfamily\footnotesize,
    breakatwhitespace=false,         
    breaklines=true,                 
    captionpos=b,                    
    keepspaces=true,                 
    numbers=left,                    
    numbersep=5pt,                  
    showspaces=false,                
    showstringspaces=false,
    showtabs=false,                  
    tabsize=2
}

\makeatletter
\let\NAT@parse\undefined
\makeatother
\usepackage[pdfa,colorlinks,bookmarksopen,bookmarksnumbered,allcolors=codegreen]{hyperref}
\usepackage[english]{babel}

\usepackage[nameinlink,capitalise]{cleveref}
\crefname{line}{line}{lines}
\crefname{figure}{Fig.}{Figs.}
\Crefname{figure}{Fig.}{Figs.}
\crefname{equation}{Eq.}{Eqs.}
\Crefname{equation}{Eq.}{Eqs.}
\crefname{section}{Sec.}{Secs.}
\Crefname{section}{Sec.}{Secs.}
\crefname{definition}{Def.}{Defs.}
\Crefname{definition}{Def.}{Defs.}
\crefname{algorithm}{Alg.}{Algs.}
\Crefname{algorithm}{Alg.}{Algs.}
\crefname{assumption}{Asm.}{Asms.}
\Crefname{assumption}{Asm.}{Asms.}
\crefname{theorem}{Thm.}{Thms.}
\Crefname{theorem}{Thm.}{Thms.}
\crefname{listing}{Lst.}{Lsts.}
\Crefname{listing}{Lst.}{Lsts.}
\crefname{table}{Tbl.}{Tbls.}
\Crefname{table}{Tbl}{Tbls.}
\crefname{subassumption}{Asm.}{Asms.}
\Crefname{subassumption}{Asm.}{Asms.}

\newcommand{\methodname}{CaStL\xspace}

\title{\LARGE \bf
CaStL: \underline{C}onstraints \underline{a}s \underline{S}pecifications \underline{t}hrough \underline{L}LM Translation for Long-Horizon Task and Motion Planning
}

\author{Weihang Guo, Zachary Kingston, and Lydia E. Kavraki%
\thanks{WG, and LEK are affiliated with the Department of Computer Science, Rice University, Houston TX, USA {\tt\small \{wg25, kavraki\}@rice.edu}. LEK is also affiliated with the Ken Kennedy Institute at Rice University. ZK is affiliated with the Department of Computer Science, Purdue University, West Lafayette IN, USA {\tt\small zkingston@purdue.edu}. This work was supported in part by
NSF RI 2336612 and Rice University funds.
}
}

\begin{document}

\maketitle
\thispagestyle{empty}
\pagestyle{empty}

\input{includes/00_Abstract}
\input{includes/01_Introduction}
\input{includes/03_Problem_Formulation}

\input{includes/02_Related_Works}

\input{includes/04_Method}

\input{includes/05_Experiment}

\input{includes/06_Conclusion}

\printbibliography{}

\end{document}

%% file: includes/00_Abstract.tex
\begin{abstract}
 Large Language Models (LLMs) have demonstrated remarkable ability in long-horizon Task and Motion Planning (TAMP) by translating clear and straightforward natural language problems into formal specifications such as the Planning Domain Definition Language (PDDL). However, real-world problems are often ambiguous and involve many complex constraints. In this paper, we introduce \underline{C}onstraints \underline{a}s \underline{S}pecifications \underline{t}hrough \underline{L}LMs (CaStL), a framework that identifies constraints such as goal conditions, action ordering, and action blocking from natural language in multiple stages. CaStL translates these constraints into PDDL and Python scripts, which are solved using an custom PDDL solver. Tested across three PDDL domains, CaStL significantly improves constraint handling and planning success rates from natural language specification in complex scenarios.
\end{abstract}

%% file: includes/01_Introduction.tex
\section{Introduction}
\begin{figure*}
    \centering
    \includegraphics[width=1\linewidth]{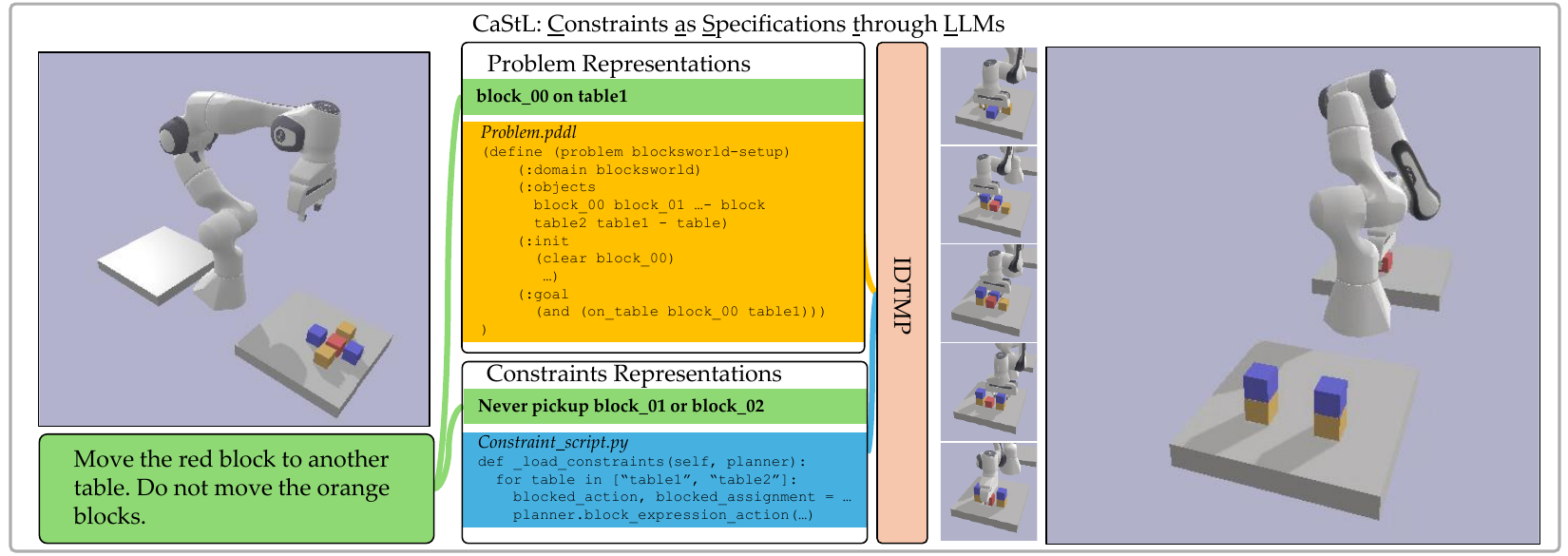}
    \caption{Our proposed method, \methodname, allows specification of Task and Motion Planning (TAMP) problems with constraints in natural language using a multi-step process (detailed in~\cref{sec:method}).
    Here, a TAMP problem (\emph{Move the red block to another table}) with an additional global constraint (\emph{Do not move the orange blocks}) is specified.
    Our approach resolves ambiguities and breaks the problem down into a PDDL specification and set of \emph{constraints} that are added to a SMT-based TAMP solver (IDTMP)~\cite{dantam2018incremental} with a Python API.
    This solver is capable of resolving motion constraints (here, the red block cannot be grasped without moving one colored pair of blocks out of the way).
    The color of each step corresponds to the module with the same color in \cref{fig:three_methods}.}
    \label{fig:one}
\end{figure*}
Task and Motion Planning (TAMP)~\cite{garrett2021integrated} methods for long-horizon robot decision-making are powerful tools capable of solving complex manipulation tasks. However, TAMP methods usually require explicit modeling of the problem domain in formal languages such as the Planning Domain Definition Language (PDDL)~\cite{fox2003pddl2}, thus burdening the system's end-user to specify desired behavior.
Recent work in Large Language Models (LLMs) has provided a means of reasoning directly from natural language (NL). Some TAMP approaches use LLMs to directly generate task sequences~\cite{wu2023tidybot, song2023llm, rana2023sayplan}. 
However, while very powerful for reasoning in general domains, LLMs have inherent limitations (e.g.,  hallucinations~\cite{zhou2023context, ji2023towards, kambhampati2024can}) that prove difficult to overcome in domains that require complex, non-monotone reasoning.

Alternatively, LLMs can be used as translators~\cite{liu2023llmp, liu2024delta} to convert an NL task into a formal language specification, which can then be solved by a standard method. These methods achieve higher success in domains when the NL task is clear and straightforward---real-world problems are often ambiguous and involve intricate \emph{constraints} on the solution. For example, a robot may need to avoid entering restricted areas in a warehouse, requiring navigation through many different rooms, or refrain from touching fragile objects while cleaning a table, thus requiring many additional, potentially non-intuitive steps to clear the table. This work focuses on problem translation in the presence of task constraints, by which we mean logical statements that are desired to hold over the state of the world or the actions taken by the system. Some methods directly address this limitation and focus on more expressive formal languages to encode constraints (e.g., AutoTAMP~\cite{chen2024autotampautoregressivetaskmotion} and DELTA~\cite{liu2024delta}). These methods also are limited and cannot be applied to more general TAMP problems or problems with different kinds of constraints, due to reliance on specific languages (e.g., STL for AutoTAMP) which do not scale to domains with many actions and objects, or by mechanism, e.g., DELTA's breaking down tasks into subtasks, which cannot encode certain types of task constraints.

In this paper, we extend the LLM-as-a-Translator framework to detect and handle constraints expressed in NL. Our approach, \textbf{CaStL} (Constraints as Specifications through LLMs) uses LLMs to identify various constraints (e.g., goal conditions, forcing orders of actions, blocking certain actions from occurring in given states, and handling extended attributes of objects) in NL in multiple stages. Problems are encoded into an SMT-based PDDL solver~\cite{de2008z3} through a combination of translation to PDDL as well as a custom PDDL modeling Python API, in which we leverage the capabilities of LLMs as Code Generators to flexibly specify complex constraints, with error, syntax, and semantic checking. Our method also addresses motion constraints, common in real-world robotics, where task plans often fail due to unreachable objects or invalid grasps by using a full TAMP stack, based on prior work in constraint-based TAMP~\cite{dantam2018incremental}.

%% file: includes/03_Problem_Formulation.tex
\section{Preliminaries}
\label{sec:prelims}
Our work addresses translating natural language (NL) tasks into specifications for Task and Motion Planning (TAMP) approaches.
We assume that we are given a model of the robot, environment, a Planning Domain Definition Language (PDDL) domain and a partially specified problem that specifies what actions and predicates are available as well as what objects are in the scene.
We denote this partially-specified PDDL problem with $(\mathcal{P}, \mathcal{A}, \mathcal{O})$, where $P(o_1, \dots, o_n) \in \mathcal{P}, o_i \in \mathcal{O}, P \rightarrow \{0, 1\}$ is the set of predicate functions and $A(o_1, \dots, o_n) \in \mathcal{A}$ is the set of actions, both grounded by object arguments.
We also include \emph{object attributes} $D(o) \in \mathcal{D}, o \in \mathcal{O}, D \rightarrow \{0, 1\}$, which encode additional properties about objects, e.g., if it is the color red, if it is heavy, and so on.
A state of the world is a set of predicates $s = \{P_1, \dots, P_n\}$ which are true, all other predicates are assumed to be false (a closed world).
Actions have preconditions (a logical expression using predicates as atoms), effects (a set of predicates that become either $\{0, 1\}$ at the next time step), and---importantly for TAMP---motion grounding that can be executed by the robot to achieve the action (e.g., finding a motion plan for a manipulator arm).
Even if an action's preconditions hold, it may not be the case that it is executable (e.g., a motion plan could not be found), thus meriting TAMP algorithms which consider the problem of not only finding a feasible task plan (a sequence of actions $\pi = A_1, \dots, A_n$), but a sequence of feasible motions that achieves the sequence of actions.

Core to any TAMP solver is the ability to consider motions failures by enumerating a number of \emph{different task plans} or attempting many motion groundings.
For efficient solving, TAMP methods impose \emph{constraints} on their task planning~\cite{dantam2018incremental, vu2024coast}, blocking actions from being taken in certain states (or under some logical expression, e.g., when objects are at specific locations), thus efficiently enumerating task plans that are more likely to be feasible.
Our approach leverages this mechanism to impose additional \emph{task constraints} on solutions, thus supporting NL translation of complex specifications.

\subsection{Task Constraints}
\label{sec:constraints}

We consider a number of different constraint classes,
all of which are built upon logical expressions $\phi$, where formulae are recursively defined with predicate atoms and operators such as \emph{and}, \emph{or}, \emph{implies}, \emph{not}, as well as universal $\forall$ and existential $\exists$ quantification over objects or object attributes.
The constraints listed below are the ones considered in this work: many other constraints are possible and easy to encode in our framework, which is general to most logical expressions.

\subsubsection{Attribute Constraints}
As described, attributes of objects can be used in constraint expressions.
While not directly supported in PDDL for quantification, our approach introduces a multi-stage approach to include attributes in constraint expressions, as described in \cref{sec:refine}.

\subsubsection{Eventual Constraints}
Eventual constraints specify that certain conditions must hold true at the end of the plan.
These constraints ensure that specific tasks are completed at some point before the plan's completion, regardless of the order in which other actions are executed, and are expressed as a conjunction of predicates in the PDDL goal.

\subsubsection{Global Constraints}
In some problems it is desirable to avoid certain conditions \emph{always}.
For example, always avoiding a certain room, never picking up a specific block (\cref{fig:one}).
Much like preconditions, these are expressed as logical expressions that apply to the state of the world, but they must always hold over every state in the task plan.

\subsubsection{Implication Constraints}
We also want to enforce sequential constraints---i.e., preventing certain actions from being taken until another action has been taken, similar to concepts in temporal logic.
These take form of a logical expression that must hold true before a specific grounding or quantification of an action can be taken, in addition to the actions normal preconditions.
For example, \emph{``blue blocks can only be picked up after the red blocks are stacked''} would be an implication constraint.

%% file: includes/02_Related_Works.tex
\section{Related Work}

Task planning involves finding a sequence of actions to transition from a given start state to a desired goal condition (e.g., STRIPS~\cite{fikes1971strips}). Many logics and languages can encode task planning problems, such as Linear Temporal Logic (LTL)~\cite{he2015towards, clarke1997model, kress2009temporal}, Signal Temporal Logic (STL)~\cite{maler2004monitoring}, context-free grammars~\cite{dantam2013motion}, and others. This work uses the Planning Domain Definition Language (PDDL)~\cite{fox2003pddl2} due to its common usage, human-readable format, and factored representation. Moreover, in PDDL 3.0~\cite{Gerevini2005PDDLConstraints}, the ability to consider \emph{constraints} on the solution was added, which enforces additional logical conditions that must hold over the found task---but this feature is poorly supported by solvers.

Having the task planner consider motion constraints is essential in TAMP solving, as many actions a robot might take might be infeasible due to geometric conditions, e.g., an object is blocking the gripper from picking up a block, and so on. Works such as IDTMP~\cite{dantam2018incremental} and COAST~\cite{vu2024coast} use constraint-aware solvers in order to focus the search and more efficiently enumerate possible solutions. In particular, Satisfiability Modulo Theories (SMT) solvers~\cite{de2008z3} used by IDTMP are a flexible approach for incorporating these constraints into TAMP.
We extend the SMT-based solving of IDTMP in this work to consider the additional constraints discussed in~\cref{sec:constraints}.

\subsection{LLMs for Task and Motion Planning}

There are two broad categories that describe how LLMs have been used to solve TAMP specified in Natural Language (NL): LLMs as task planners, and LLMs as translators.

First, LLMs-as-task-planners use the reasoning capabilities (e.g.,~\citet{huang2022language}) of LLMs directly to generate a task plan, either step-by-step or as a whole.
Initial works~\cite{wu2023embodied, wu2023tidybot, vemprala2024chatgpt} use zero-shot generation of action sequences from an NL description, but they suffer from poor execution success;
later methods either generate a new plan upon failure~\cite{song2023llm, wang2023describe, shinn2024reflexion, raman2022cape} or iteratively find the next action to execute~\cite{huang2022inner, yoneda2023statler, fu2024scene, zhang2024fltrnn}.
To further ground the actions (i.e., finding feasible motions to execute), some works~\cite{brohan2023can, lin2023text2motion, shah2023navigation, paulius2023long} use affordance functions or other heuristics to guide LLM inference, learn downstream networks to use LLM output~\cite{zhang2024fltrnn}, or integrate environment data into planning~\cite{xiang2024language, rana2023sayplan, fu2024scene}.
These methods perform very well on tasks that require ``common-sense'' reasoning and have many independent actions, but fail to scale to more complex problems, due to limitations such as context-faithfulness~\cite{zhou2023context}, hallucination~\cite{ji2023towards}, and principle reasoning~\cite{kambhampati2024can}.
In this work, we are concerned with tasks with many \emph{constraints} on valid actions and may require complex, non-monotone reasoning, to which LLMs-as-task-planners are ill-suited.

To overcome the natural limitations of LLM reasoning, many approaches instead use the LLM as a \emph{translator}, to convert NL requests into a formal language, for example, LTL~\cite{fuggitti2023nl2ltl, pan2023data}, STL~\cite{chen2023nl2tl, chen2024autotampautoregressivetaskmotion}, and MTL~\cite{manas2024tr2mtl}.
Relevant to this work, many approaches have translated NL into PDDL~\cite{liu2024delta, arora2024anticipate, birr2024autogpt+p, liu2023llmp, wang2024llm} and have even generated the PDDL domain~\cite{shirai2023vision, han2024interpret}.
However, none of PDDL translation approaches have \emph{directly} addressed translation of NL tasks that include constraints in addition to reaching the goal.
One approach, DELTA~\cite{liu2024delta}, splits the NL task into subgoals, handling some constraints that require actions done in a certain order.
However, this approach does not guarantee optimal makespan, and was not designed to handle complex constraints---performance remains low on many relevant problems (see \cref{sec:validation}).
Some approaches which translate NL to temporal logics (e.g., NL2TL~\cite{chen2023nl2tl} and AutoTAMP~\cite{chen2024autotampautoregressivetaskmotion}) directly handle ordering constraints due to the nature of the formal language.
However, they demonstrate results only in 2D domains and consider just two actions: \emph{enter} and \emph{not-enter}.
While not able to handle the full gamut of temporal constraints, our approach handles implication and global constraints (\cref{sec:constraints}) in PDDL domains with 3D manipulation workspaces.

%% file: includes/04_Method.tex
\section{\methodname}
\label{sec:method}
\begin{figure}
    \centering
    \includegraphics[width=1\linewidth]{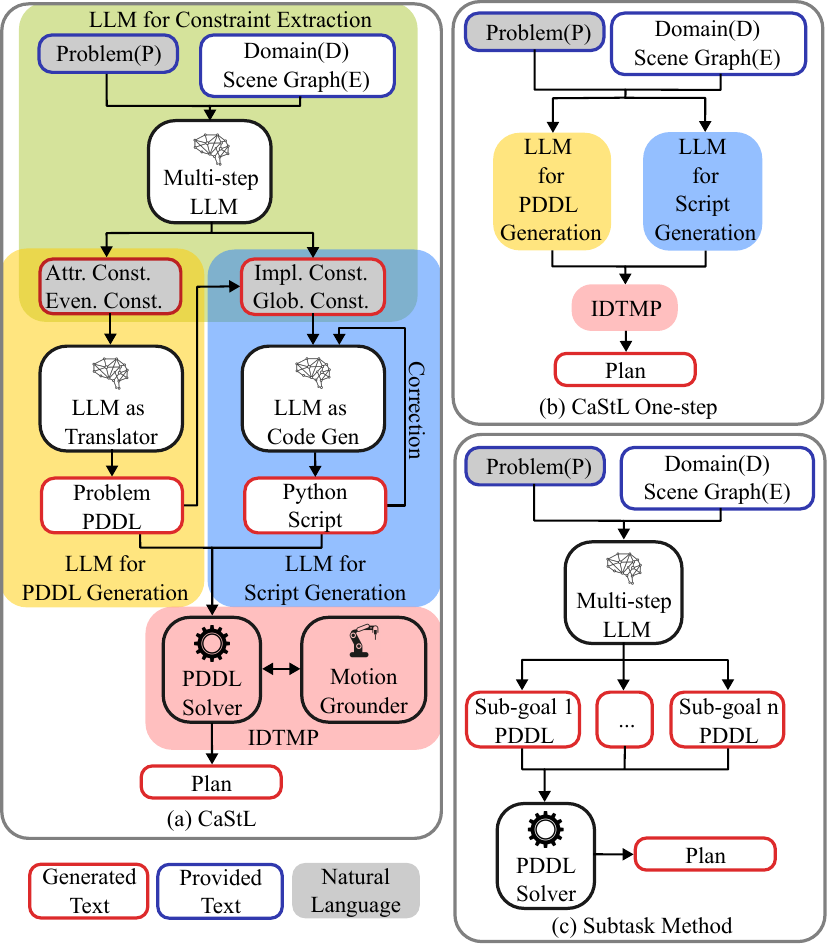}
    \caption{Illustrations of approaches.
(a) In our approach, \methodname, constraints are first extracted through multi-step LLM queries (\cref{sec:refine}).
Then, the LLM translates natural language constraints into PDDL (\cref{sec:translate_pddl}) as well as Python scripts which use an API on our SMT-based PDDL solver within a TAMP algorithm, IDTMP~\cite{dantam2018incremental} (\cref{sec:constraint_script}).
(b) \methodname One-step is an ablation of \methodname, without the multi-step LLM process for extracting constraints.
(c) Baseline.
The problem is decomposed into natural language subproblems, which are translated sequential into PDDL.
}
    \label{fig:three_methods}
\end{figure}

The goal of our method, \methodname (Constraints as Specifications through LLMs), is to parse natural language (NL) tasks in a given Planning Domain Definition Language (PDDL) domain into specifications understood by a TAMP solver to find a feasible TAMP plan to execute on a system.
Our approach focuses on specifications as a set of \emph{constraints} (detailed in~\cref{sec:prelims}), that involve not only a desired end goal condition (\emph{eventual constraints}), but also additional constraints that limit what actions can be taken at a given step in the plan, e.g., never taking certain actions (\emph{global constraints}) or only taking an action once another condition has been fulfilled (\emph{implication constraints}).

To achieve this, we first perform a multi-step query to an LLM (\cref{sec:refine}).
We provide the PDDL domain, an environment scene graph, and the user's NL query as inputs to the LLM, which translates the NL query into a list of constraints.
Next, the LLM translates the \emph{eventual constraints} into a PDDL problem (\cref{sec:translate_pddl}).
The \emph{global} and \emph{implication constraints} are converted by multi-step LLM prompt with semantic, syntactic, and error checking into a Python script which adds the constraints to the TAMP solver (\cref{sec:constraint_script}).
The fully specified problem is then solved by a TAMP planner, considering motion feasibility, and a final plan is generated if possible within the time limit.

\subsection{LLM for Multi-step Constraint Extraction}\label{sec:refine}
The LLM takes a user-specified problem in NL, a PDDL domain, and an environment scene graph as input.
In the first query, the LLM resolves ambiguities by matching pronouns and \emph{attribute constraints} to the environment.
For example, the original problem, \emph{``You must visit the rooms that have a bed. But, you should visit the one with the largest bed first."} becomes \emph{``The robot must visit room02, room05, and room06. The robot should visit room05 first."}
In this example \emph{``rooms with a bed"} is an attribute constraint. In the second query, after resolving ambiguities, the LLM determines if the NL problem expresses implication and global constraints. If those constraints are present, the third query identifies eventual, implication, and global constraints, guided by in-context examples. Otherwise, only eventual constraints are identified.
In this case, \emph{``The robot must visit room02, room05, room06."} is an eventual constraint, and \emph{``The robot should visit room05 first."} is an implication constraint.

\subsection{LLM for PDDL Problem Generation}\label{sec:translate_pddl}
We employ the same approach as LLM+P~\cite{liu2023llmp} for PDDL generation.
The input consists of all eventual constraints in NL, a PDDL domain, the environment, and in-context examples in the same domain.
The key difference between \methodname and LLM+P in this module is that we use the output of the prior multi-step query to disambiguate the problem as input.
We include a \textit{one-step} ablation, which does not have the above multi-step process in our experiments (\cref{sec:alg_ablation}) to show that adding this step significantly increases success.

\subsection{LLM for Constraint Script Generation and Correction}\label{sec:constraint_script}

\begin{lstlisting}[language=Python, caption={Python script for constraint \textit{visiting all rooms before the backyard}. This script is automatically generated by GPT-4o.}, label={lst:python_script},float]
def _load_constraints(self, planner):
    backyard_unvisited = self.make_grounded_predicate("visited", ["robot1", "backyard"])
    
    # Create constraints for all the rooms except backyard
    rooms = ["kitchen", "bedroom1", "bedroom2", "restroom"]
    constraints = []
    for room in rooms:
        condition = self.make_grounded_predicate("visited", ["robot1", room])
        constraints.append(condition)
        
    and_condition = pd.make_and(constraints)
    
    # Block the `move` action to the backyard if not all other rooms are visited
    blocked_action, blocked_assignment = self.make_action_assignment("move", ["robot1", "living-room", "backyard"])
    planner.block_expression_action(blocked_action, blocked_assignment, pd.make_not(and_condition), pd.Assignment())
\end{lstlisting}

Although PDDL supports constraint specification since 3.0~\cite{Gerevini2005PDDLConstraints}, it is rarely used and thus not present in LLM training data or well supported.
Thus, the success rate for LLMs to directly translate constraints into PDDL (i.e., the \texttt{:constraints} field) is low.
Therefore, we instead prompt the LLM to extract and translate constraints into a Python script that uses a custom API to build logical expressions and add constraints to the task planner.

The translation process is achieved in multiple steps: first, each additional constraint in natural language (NL) is paraphrased into a constraint-specific format.
Implication constraints are paraphrased into \texttt{Never/Always <expression>} and global constraints are paraphrased into \texttt{Do not <action> until <expression> is/is not true}.
Next, the paraphrased expressions are translated into a Python script by the LLM (an example is shown in~\cref{lst:python_script}).
Syntax errors are handled with corrective re-prompting~\cite{raman2022reprompt}, where the error message is given to the LLM to regenerate the script.
We also use the LLM to evaluate whether the generated script was semantically consistent with the original instructions.

We use the SMT-based~\cite{de2008z3} TAMP algorithm from \citet{dantam2018incremental}, where the solver maintains a constraint stack to generate alternate task plans with an increasing horizon.
In addition to internally generated motion constraints, we provide a Python API for scripting additional constraints onto the constraint stack, examples of which are visible in~\cref{lst:python_script}.

An ablation of this module is evaluated in~\cref{sec:var_script}, where instead of having the LLM generate Python, we ask it to translate the constraints into a JSON schema (\cref{lst:json}), which is then parsed into the solver.
The results show that JSON translation performs worse than scripting at accurately modeling constraints.

%% file: includes/05_Experiment.tex
\section{Validation}
\label{sec:validation}

\input{figs/results/big_table}

We evaluate on three domains: \textsc{HouseChip} (\textsc{HC}), \textsc{Kitchen} (\textsc{KT}), and \textsc{BlocksWorld} (\textsc{BW}), described further in~\cref{sec:domain_desc}.
For each domain, we consider two classes of randomly generated environments, one simple (ending with \textsc{1}) and one complex (ending with \textsc{2}), with complexity based on the number of objects.
We also consider increasing complexity of tasks: only eventual constraints (\textsc{No}), eventual and implication constraints (\textsc{Impl}), eventual and global constraints (\textsc{Glob}), eventual, implication, and global constraints (\textsc{Impl Glob}), finally all types of constraints (\textsc{Impl Glob Attr}).
11 trials of each task in each environment are evaluated for each method, discussed in~\cref{sec:alg_ablation}.
We use GPT-4o for all our experiments due to its balance between cost and performance.

\begin{figure}
    \centering
    \begin{subfigure}{0.55\linewidth}
        \centering
        \includegraphics[width=\linewidth]{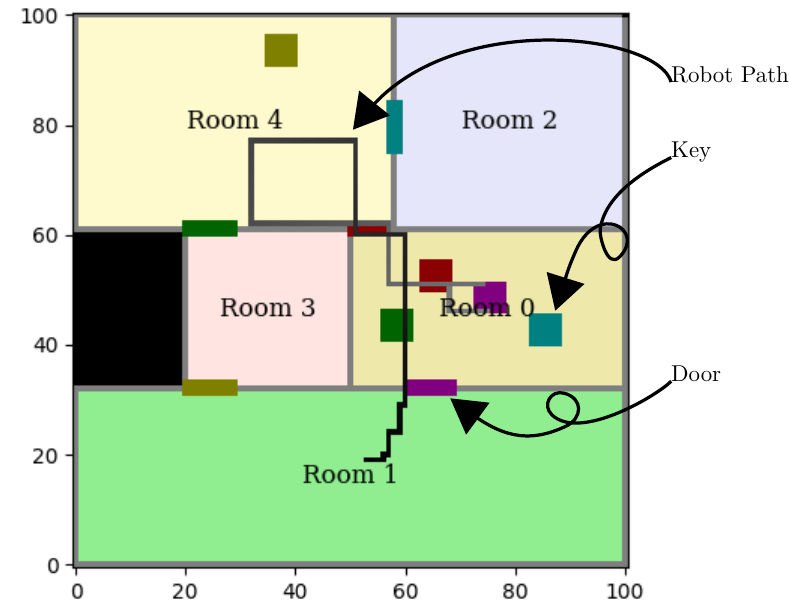}
        \caption{\textsc{HouseChip(HC)}}
        \label{fig:hc_domain}
    \end{subfigure}
    \begin{subfigure}{0.4\linewidth}
        \centering
        \includegraphics[width=\linewidth]{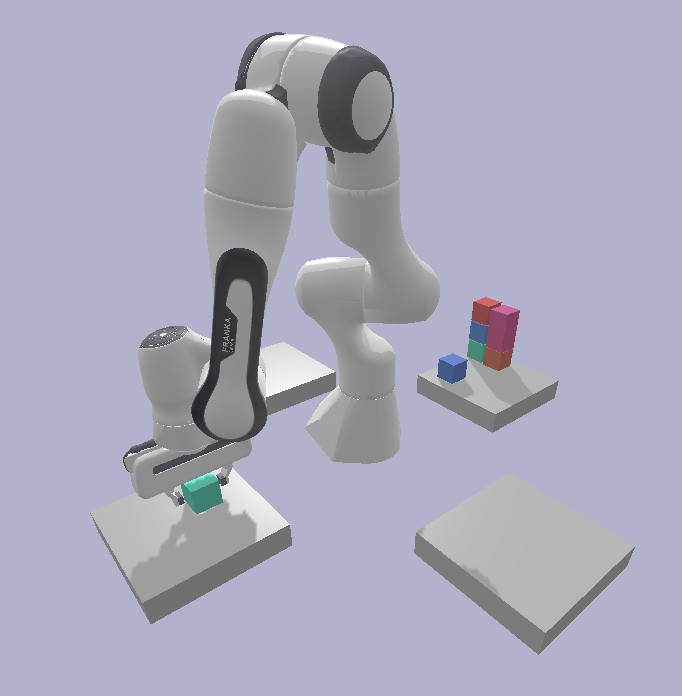}
        \caption{\textsc{BlocksWorld(BW)}}
        \label{fig:bw_domain}
    \end{subfigure}
    \vspace{1em}
    \caption{(a) The \textsc{HC} domain features a robot starting in Room 0, tasked with visiting a list of rooms, each of which locked by a corresponding key.
      (b) The \textsc{BW} domain, which consists of pick, place, stack, and unstack actions with a number of blocks and tables.
    }
    \label{fig:domains}
\end{figure}

\subsection{Domain Descriptions}\label{sec:domain_desc}
The \textbf{\textsc{HouseChip} (\textsc{HC})} domain (\cref{fig:hc_domain}) is inspired by the HouseWorld and Chip’s Challenge domains from AutoTAMP~\cite{chen2024autotampautoregressivetaskmotion}.
Despite their similarity, adapting AutoTAMP to this domain for a fair comparison is non-trivial.
AutoTAMP simplifies doors, keys, walls, and rooms into a single \emph{room} entity, focusing on two actions: \emph{enter} or \emph{not enter}. Additionally, we found that some parameters in their algorithm are sensitive to the specific setup. Thus, we excluded AutoTAMP from the comparison.
Eventual constraints are to visit rooms, with all doors initially locked and can be unlocked by obtaining the corresponding keys.
The robot must navigate within 5 units of the room’s center to mark it as visited; $A^*$~\cite{astar} is used to ground motions.
Implication constraints specify room visitation order.
Global constraints restrict room access or key collection.
This domain is custom-designed, so LLMs lack prior training data.

The \textbf{\textsc{Kitchen} (\textsc{KT})} domain is adapted from the 2014 International Planning Competition~\cite{vallati20152014}.
Eventual constraints task the robot with making and delivering sandwiches to some or all children.
Some children are allergic to gluten, requiring both normal and gluten-free sandwiches.
Implication constraints force delivery order or force preparation of all sandwiches before serving.
Global constraints block certain ingredients or trays.
LLMs likely have limited training data for this domain.

In \textbf{\textsc{BlocksWorld} (\textsc{BW})}, a manipulator arranges blocks in a specified order using the Franka Emika Panda robot, with RRT-Connect~\cite{kuffner00_rrtconnect} to ground motion.
Implication constraints prevent picking up certain blocks until other criteria are met.
Global constraints prohibit moving certain blocks or placing specific blocks on specific tables.
LLMs likely have ample training data for this domain.
Block positions are randomized in each trial, and thus the motion planner might succeed in one instance but fail in another.

\subsection{Algorithm Ablations and Baseline}\label{sec:alg_ablation}

We present an ablation of \methodname (shown in \cref{fig:three_methods}b), where LLMs directly translate into PDDL and Python script, bypassing multi-step constraint extraction (\methodname One-step).
We also implemented a baseline approach following the DELTA~\cite{liu2024delta} architecture (\cref{fig:three_methods}c) which decomposes the problem into sub-problems and sequentially solves generated PDDL problems (Subtask).
Since DELTA does not provide their full prompt, we created our own to achieve problem decomposition.
We use the same prompt and in-context examples as \methodname for translation.
While DELTA includes PDDL domain generation via LLMs, we excluded this as it is not the focus of this paper.

\subsection{General Results}\label{sec:general_result}

Results are shown in \cref{tab:big_table}.
Two success rates are reported: one for pure task planning, one for task and motion planning.
We also record the average number of input tokens and the time taken by the LLM per run.
The TAMP solver is given a timeout of 60 seconds.
Success is evaluated with a script to compare against ground truth constraints.

The full \methodname method outperforms the one-step approach.
The one-step method often reports false positives on problems without constraints, demonstrating the utility of the multi-step prompting strategy.
The total time spent on LLM queries is comparable to the one-step method, even though \methodname uses more queries due to its multi-step process.
This is likely because each step is simpler, allowing the LLM to process them efficiently.
While the baseline approach can handle some implication constraints, it struggles with global constraints, and has longer LLM processing times due to individual translation of sub-problems.

\subsection{Variants for Constraint Script Generation}\label{sec:var_script}
\input{figs/results/constraint_comp}

We compare an ablation of our Python script generation for implication and global constraint specification (e.g., \cref{lst:python_script}) to an approach where constraints are translated and parsed to and from a JSON schema (\cref{lst:json}).
We evaluate on the \textsc{BlocksWorld} domain, both with and without correction.
The results, shown in \cref{tab:constraint_comparison}, compare the \textsc{BW 3} setup, which shares the same environment as \textsc{BW 2} but includes additional constraints.
These constraints included conditions like \emph{``the robot cannot move blocks 1, 2, and 3 when block 4 is on block 5''} or \emph{``all blocks can only be placed on their original table.''}

We found that NL action names often lack one-to-one mappings (e.g., ``touch'' maps to both \textit{pick-up} and \textit{unstack} in \textsc{BW}).
NL also tends to omit indirect objects and use quantification, causing a single sentence to map to many constraints.
The Python script approach efficiently handles this with loops, while JSON requires all constraints to be explicitly specified.
For small numbers of constraints, JSON performs well, sometimes surpassing the Python script approach without correction.
However, as the number of constraints increase, LLMs struggle to capture all of them, leading to a significant drop in success rate.

\begin{lstlisting}[caption={Return implication and global constraints in JSON}, label={lst:json}, float]
[
  {
    "type": "implication",
    "action": ["pick-up", "block0", "table0"],
    "condition": [["on", "block4", "block5"]]
  },
  {
    "type": "global",
    "condition": [["not", "on_table", "block0", "table1"]]
  },
  ...
]
\end{lstlisting}

\subsection{Cross-domain Generalizability}\label{sec:cross_domain}
\input{figs/results/cross_domain_comp}

\methodname generalizes effectively across different domains when provided with in-context examples.
We further evaluate performance by in-context examples from a different domains, shown in \cref{tab:cross_comparison}, where \methodname solves the same problems in \textsc{Kitchen} using in-context examples from the \textsc{BlocksWorld} domain.
We observed that the algorithm handles implication constraints more robustly than global constraints.
Notably, for implication constraints in \textsc{KT 2}, the success rate is higher.
Failure cases fall into two categories: First, LLMs occasionally miss action parameters or reference non-existent objects; second, they misinterpret whether a condition or its negation should be applied.

%% file: figs/results/big_table.tex
\small
\begin{table*}[t]%
\setlength{\tabcolsep}{5pt} 
\begin{tabularx}{\textwidth}{c|X|c|c|c|c|c|c|c|}%

\multicolumn{1}{c|}{Cases}&\multicolumn{1}{X|}{Methods}&\multicolumn{5}{c|}{Problems and Constraints}&\multicolumn{1}{c|}{Average}&\multicolumn{1}{c|}{Time}\\ 

\multicolumn{1}{c|}{}&\multicolumn{1}{X|}{}&No&Impl&Glob&Impl Glob&Impl Glob Attr&Input&in\\
\multicolumn{1}{c|}{}&\multicolumn{1}{X|}{}&Logic / Motion&Logic / Motion&Logic / Motion&Logic / Motion&Logic / Motion&Tokens&LLM~(s)\\

\hline 
\hline
\multirow{3}{*}{HC 1}&Subtask & 91\%/91\% & 91\%/91\% & 73\%/73\% & 27\%/27\% & 18\%/18\% & 9140 & 36.13\\&\cellcolor{lightgrey}{\methodname (one step)} & \cellcolor{lightgrey}{\textbf{100\%}/\textbf{100\%}} & \cellcolor{lightgrey}{91\%/91\%} & \cellcolor{lightgrey}{91\%/91\%} & \cellcolor{lightgrey}{27\%/27\%} & \cellcolor{lightgrey}{-/-} & \cellcolor{lightgrey}{10415} & \cellcolor{lightgrey}{11.35}\\&\cellcolor{grey}{\methodname} & \cellcolor{grey}{\textbf{100\%}/\textbf{100\%}} & \cellcolor{grey}{\textbf{100\%}/\textbf{100\%}} & \cellcolor{grey}{\textbf{100\%}/\textbf{100\%}} & \cellcolor{grey}{\textbf{73\%}/\textbf{73\%}} & \cellcolor{grey}{\textbf{73\%}/\textbf{73\%}}& \cellcolor{grey}{15112} & \cellcolor{grey}{10.79}\\\hline\multirow{3}{*}{HC 2}&Subtask & 82\%/82\% & 91\%/91\% & 18\%/18\% & 82\%/82\% & \textbf{91\%}/\textbf{91\%} & 7767 & 17.60\\&\cellcolor{lightgrey}{\methodname (one step)} & \cellcolor{lightgrey}{91\%/91\%} & \cellcolor{lightgrey}{\textbf{100\%}/\textbf{100\%}} & \cellcolor{lightgrey}{91\%/91\%} & \cellcolor{lightgrey}{\textbf{91\%}/\textbf{91\%}} & \cellcolor{lightgrey}{18\%/18\%} & \cellcolor{lightgrey}{7321} & \cellcolor{lightgrey}{8.24}\\&\cellcolor{grey}{\methodname} & \cellcolor{grey}{\textbf{100\%}/\textbf{100\%}} & \cellcolor{grey}{\textbf{100\%}/\textbf{100\%}} & \cellcolor{grey}{\textbf{100\%}/\textbf{100\%}} & \cellcolor{grey}{\textbf{91\%}/\textbf{91\%}} & \cellcolor{grey}{\textbf{91\%}/\textbf{91\%}}& \cellcolor{grey}{9195}& \cellcolor{grey}{9.42}\\

\hline\multirow{3}{*}{KT 1}&Subtask & 55\%/- & 9\%/- & -/- & -/- & -/- & 9140 & 36.13\\&\cellcolor{lightgrey}{\methodname (one step)} & \cellcolor{lightgrey}{36\%/-} & \cellcolor{lightgrey}{\textbf{100\%}/-} & \cellcolor{lightgrey}{55\%/-} & \cellcolor{lightgrey}{91\%/-} & \cellcolor{lightgrey}{\textbf{100\%}/-} & \cellcolor{lightgrey}{10415} & \cellcolor{lightgrey}{11.35}\\&\cellcolor{grey}{\methodname} & \cellcolor{grey}{\textbf{91\%}/-} & \cellcolor{grey}{\textbf{100\%}/-} & \cellcolor{grey}{\textbf{82\%}/-} & \cellcolor{grey}{\textbf{100\%}/-} & \cellcolor{grey}{91\%/-}& \cellcolor{grey}{15040} & \cellcolor{grey}{11.44}\\\hline\multirow{3}{*}{KT 2}&Subtask & 9\%/- & 45\%/- & -/- & -/- & -/- & 11448 & 45.71\\&\cellcolor{lightgrey}{\methodname (one step)} & \cellcolor{lightgrey}{55\%/-} & \cellcolor{lightgrey}{73\%/-} & \cellcolor{lightgrey}{45\%/-} & \cellcolor{lightgrey}{\textbf{64\%}/-} & \cellcolor{lightgrey}{27\%/-} & \cellcolor{lightgrey}{10711} & \cellcolor{lightgrey}{10.86}\\&\cellcolor{grey}{\methodname} & \cellcolor{grey}{\textbf{100\%}/-} & \cellcolor{grey}{\textbf{100\%}/-} & \cellcolor{grey}{\textbf{82\%}/-} & \cellcolor{grey}{\textbf{64\%}/-} & \cellcolor{grey}{\textbf{64\%}/-}& \cellcolor{grey}{15938} & \cellcolor{grey}{11.76}\\

\hline\multirow{3}{*}{BW 1}&Subtask & -/- & 9\%/- & -/- & -/- & -/- & - & -\\&\cellcolor{lightgrey}{\methodname (one step)} & \cellcolor{lightgrey}{82\%/45\%} & \cellcolor{lightgrey}{55\%/55\%} & \cellcolor{lightgrey}{64\%/36\%} & \cellcolor{lightgrey}{36\%/55\%} & \cellcolor{lightgrey}{36\%/36\%} & \cellcolor{lightgrey}{4520} & \cellcolor{lightgrey}{8.11}\\&\cellcolor{grey}{\methodname} & \cellcolor{grey}{\textbf{100\%}/\textbf{64\%}} & \cellcolor{grey}{\textbf{64\%}/\textbf{64\%}} & \cellcolor{grey}{\textbf{100\%}/\textbf{73\%}} & \cellcolor{grey}{\textbf{73\%}/\textbf{73\%}} & \cellcolor{grey}{\textbf{64\%}/\textbf{73\%}}& \cellcolor{grey}{7424} & \cellcolor{grey}{10.52}\\\hline\multirow{3}{*}{BW 2}&Subtask & 55\%/- & 27\%/- & 27\%/- & 9\%/- & 27\%/- & 4124 & 12.32\\&\cellcolor{lightgrey}{\methodname (one step)} & \cellcolor{lightgrey}{64\%/55\%} & \cellcolor{lightgrey}{82\%/55\%} & \cellcolor{lightgrey}{18\%/18\%} & \cellcolor{lightgrey}{\textbf{64\%}/\textbf{55\%}} & \cellcolor{lightgrey}{\textbf{64\%}/36\%} & \cellcolor{lightgrey}{7681} & \cellcolor{lightgrey}{15.80}\\&\cellcolor{grey}{\methodname} & \cellcolor{grey}{\textbf{82\%}/\textbf{82\%}} & \cellcolor{grey}{\textbf{100\%}/\textbf{64\%}} & \cellcolor{grey}{\textbf{55\%}/\textbf{45\%}} & \cellcolor{grey}{\textbf{64\%}/45\%} & \cellcolor{grey}{\textbf{64\%}/\textbf{45\%}}& \cellcolor{grey}{8021} & \cellcolor{grey}{16.41}\\
\end{tabularx}%
\caption{Success rates for pure task planning (\emph{Logic}) and after grounding with motion planning (\emph{Motion}) are presented for the \textsc{BW}, \textsc{HC}, and \textsc{KT} domains.
The \textsc{KT} domain and the Subtask method do not ground motions, denoted by '-'.
A success rate of 0\% is also marked as ‘-’.
\emph{Average input tokens} is the average total number of tokens spent on each trial with task planning success, average time on LLM queries given by \emph{Time in LLM}.
\emph{No, Impl, Glob}, and \emph{Attr} refer respectively to problems with no, implication, global, and attribute constraints.
Constraints are added to the same initial \emph{No} problem for each case.
All experiments use GPT-4o.
}
\label{tab:big_table}%
\end{table*}
\normalsize

%% file: figs/results/constraint_comp.tex
\small
\begin{table}[t]%
\setlength{\tabcolsep}{5pt}

\begin{tabularx}{\linewidth}{c|c|X|c}%

Cases & Problems & Methods & Success Rate\\
\hline \hline\multirow{12}{*}{BW 1} &\multirow{3}{*}{Glob}& Script & \textbf{100\%} \\ & & \cellcolor{lightgrey}{Script w/o correction} & \cellcolor{lightgrey}{91\%} \\ & & \cellcolor{grey}{JSON}  & \cellcolor{grey}{\textbf{100\%}} \\\cline{2-4}&\multirow{3}{*}{Impl}& Script & \textbf{64\%} \\ & & \cellcolor{lightgrey}{Script w/o correction} & \cellcolor{lightgrey}{\textbf{64\%}} \\ & & \cellcolor{grey}{JSON}  & \cellcolor{grey}{\textbf{64\%}} \\\cline{2-4}&\multirow{3}{*}{Impl Glob}& Script & \textbf{73\%} \\ & & \cellcolor{lightgrey}{Script w/o correction} & \cellcolor{lightgrey}{64\%} \\ & & \cellcolor{grey}{JSON}  & \cellcolor{grey}{\textbf{73\%}} \\\cline{2-4}&\multirow{3}{*}{Impl Glob Attr}& Script & \textbf{64\%} \\ & & \cellcolor{lightgrey}{Script w/o correction} & \cellcolor{lightgrey}{45\%} \\ & & \cellcolor{grey}{JSON}  & \cellcolor{grey}{\textbf{64\%}} \\\hline\multirow{12}{*}{BW 3} &\multirow{3}{*}{Glob}& Script & \textbf{36\%} \\ & & \cellcolor{lightgrey}{Script w/o correction} & \cellcolor{lightgrey}{18\%} \\ & & \cellcolor{grey}{JSON}  & \cellcolor{grey}{-} \\\cline{2-4}&\multirow{3}{*}{Impl}& Script & \textbf{91\%} \\ & & \cellcolor{lightgrey}{Script w/o correction} & \cellcolor{lightgrey}{\textbf{91\%}} \\ & & \cellcolor{grey}{JSON}  & \cellcolor{grey}{73\%} \\\cline{2-4}&\multirow{3}{*}{Impl Glob}& Script & \textbf{45\%} \\ & & \cellcolor{lightgrey}{Script w/o correction} & \cellcolor{lightgrey}{36\%} \\ & & \cellcolor{grey}{JSON}  & \cellcolor{grey}{-} \\\cline{2-4}&\multirow{3}{*}{Impl Glob Attr}& Script & \textbf{45\%} \\ & & \cellcolor{lightgrey}{Script w/o correction} & \cellcolor{lightgrey}{\textbf{45\%}} \\ & & \cellcolor{grey}{JSON}  & \cellcolor{grey}{-} \\\hline
\end{tabularx}%
\caption{Comparison of the Python \emph{Script} and JSON approaches for representing constraints on task planning success.
Besides this component, the rest of the \methodname method is kept the same.
\emph{Impl}, \emph{Glob}, and \emph{Attr} refer to problems expressing implication, global, and attribute constraints.
The \emph{Script} method handles natural language ambiguity and many-to-one mappings with loops, while the JSON approach requires explicit specification of all constraints.
}
\label{tab:constraint_comparison}
\end{table}
\normalsize

%% file: figs/results/cross_domain_comp.tex
\small
\begin{table}[t]%
\setlength{\tabcolsep}{5pt}

\begin{tabularx}{\linewidth}{c|c|X|c}%

Cases & Problems & Methods & Success Rate\\
\hline \hline\multirow{8}{*}{KT 1} &\multirow{2}{*}{Glob}& Same context & \textbf{91\%} \\ & & \cellcolor{grey}{BW context} & \cellcolor{grey}{18\%} \\\cline{2-4}&\multirow{2}{*}{Impl}& Same context & \textbf{100\%} \\ & & \cellcolor{grey}{BW context} & \cellcolor{grey}{73\%} \\\cline{2-4}&\multirow{2}{*}{Impl Glob}& Same context & \textbf{100\%} \\ & & \cellcolor{grey}{BW context} & \cellcolor{grey}{27\%} \\\cline{2-4}&\multirow{2}{*}{Impl Glob Attr}& Same context & \textbf{100\%} \\ & & \cellcolor{grey}{BW context} & \cellcolor{grey}{18\%} \\\hline\multirow{8}{*}{KT 2} &\multirow{2}{*}{Glob}& Same context & \textbf{82\%} \\ & & \cellcolor{grey}{BW context} & \cellcolor{grey}{27\%} \\\cline{2-4}&\multirow{2}{*}{Impl}& Same context & 73\% \\ & & \cellcolor{grey}{BW context} & \cellcolor{grey}{\textbf{82\%}} \\\cline{2-4}&\multirow{2}{*}{Impl Glob}& Same context & \textbf{73\%} \\ & & \cellcolor{grey}{BW context} & \cellcolor{grey}{9\%} \\\cline{2-4}&\multirow{2}{*}{Impl Glob Attr}& Same context & \textbf{65\%} \\ & & \cellcolor{grey}{BW context} & \cellcolor{grey}{-} \\\hline
\end{tabularx}%
\caption{Comparison of same-domain in-context examples and cross-domain in-context examples for task planning.
\emph{Impl}, \emph{Glob}, and \emph{Attr} refer to natural language problems expressing implication, global, and attribute constraints.
}
\label{tab:cross_comparison}
\end{table}

\normalsize

%% file: includes/06_Conclusion.tex
\section{Conclusion}

We present \methodname, a method which efficiently uses multi-step queries to LLMs with few-shot in-context examples to translate attribute, eventual, implication, and global constraints from natural language into a formal specification for Task and Motion Planning (TAMP).
In comparison to ablations of \methodname without multi-step querying and a baseline from the literature~\cite{liu2024delta}, we demonstrate the improved capability of our constraint-based approach in specifying complex tasks to TAMP solvers.

We also identified a few limitations.
First, the prompt and in-context examples are crucial; while they can be reused within the same domain, designing them requires expertise.
We aim to explore methods that do not rely on in-context examples or use examples that generalize across different domains.
Second, the response time and computational cost remain high: investigating local or cheaper models is of interest.
Third, we plan to support additional types of constraints, such as temporal and geometric constraints, and connect our approach to other task planning systems which may consider uncertainty and open worlds.